\documentclass[letterpaper, 10 pt, conference]{IEEEtran}

\IEEEoverridecommandlockouts                              % This command is only needed if 
\usepackage{hyperref}
\usepackage[numbers]{natbib}
\usepackage{amsmath,amsthm,amssymb}

\usepackage{siunitx}
\usepackage{leftindex}
\usepackage{bm}
\usepackage{algorithmicx} % for pseudocode
\usepackage{algorithm}% http://ctan.org/pkg/algorithms
\usepackage[noend]{algpseudocode}% http://ctan.org/pkg/algorithmicx
\usepackage{caption}
\usepackage{subcaption}
\usepackage{graphicx}
\usepackage{cuted}
\usepackage{capt-of}
\usepackage{gensymb} % for degree symbol
\usepackage{multirow}
\usepackage{booktabs}
\usepackage{bbm}

\usepackage{makecell}
\usepackage{array}

\newif\ifproofread

\newcommand{\bs}[1]{\boldsymbol{#1}}
\algrenewcommand\algorithmicrequire{\textbf{Input:}}
\algrenewcommand\algorithmicensure{\textbf{Output:}}

\usepackage[font=small,labelfont=bf]{caption}

\newif\ifanonym
\anonymfalse % for camera-ready

\ifanonym
  \hypersetup{
    pdfauthor={},
    pdftitle={},
    pdfsubject={},
    pdfkeywords={},
    pdfcreator={},
    pdfproducer={}
  }
\fi

\newcommand{\projecturl}{}
\ifanonym
  \renewcommand{\projecturl}{}
\else
  \renewcommand{\projecturl}{For more details, please visit our website: \url{https://nvlabs.github.io/CHIP/}}
\fi

\ifanonym
  \usepackage[disable]{todonotes}
\else
  \usepackage{todonotes}
\fi

\ifanonym

\fi

\usepackage{xcolor}
\usepackage[normalem]{ulem} % provides \sout, keeps \emph italic
\usepackage[final,authormarkup=none]{changes} % switch draft<->final

\setaddedmarkup{\textcolor{green!50!black}{#1}}
\setdeletedmarkup{\textcolor{red}{\sout{#1}}}
\setcommentmarkup{\textcolor{orange!80!black}{(#1)}}
\title{\LARGE \bf CHIP: Adaptive Compliance for Humanoid Control through Hindsight Perturbation}

\ifanonym
  \author{Anonymous Authors}
\else
  \author{Sirui Chen$^{1,2,\dagger}$, Zi-ang Cao$^{1,3,\dagger}$, Zhengyi Luo$^{1}$, Fernando Castañeda$^{1}$, Chenran Li$^{1}$\\
  Tingwu Wang$^{1}$, Ye Yuan$^{1}$, Linxi ``Jim" Fan$^{1}$, C. Karen Liu$^{2,\ddagger}$, Yuke Zhu$^{1,3,\ddagger}$\\
  \small$^\dagger$ Co-First Authors \space \small$^\ddagger$ Equal Advising\\
  $^{1}$NVIDIA\quad $^{2}$Stanford University \quad $^{3}$UT Austin
  \thanks{This work was done during their internships of Sirui Chen and Zi-Ang Cao at NVIDIA GEAR Lab.}

  \vspace{-40pt}
  }
\fi

\begin{document}

% Set this to false to change the new text to black
\proofreadfalse

\maketitle
\thispagestyle{empty}
\pagestyle{empty}

%%%%%%%%%%%%%%%%%%%%%%%%%%%%%%%%%%%%%%%%%%%%%%%%%%%%%%%%%%%%%%%%%%%%%%%%%%%%%%%%
\begin{strip}
  \centering
  \includegraphics[width=\linewidth]{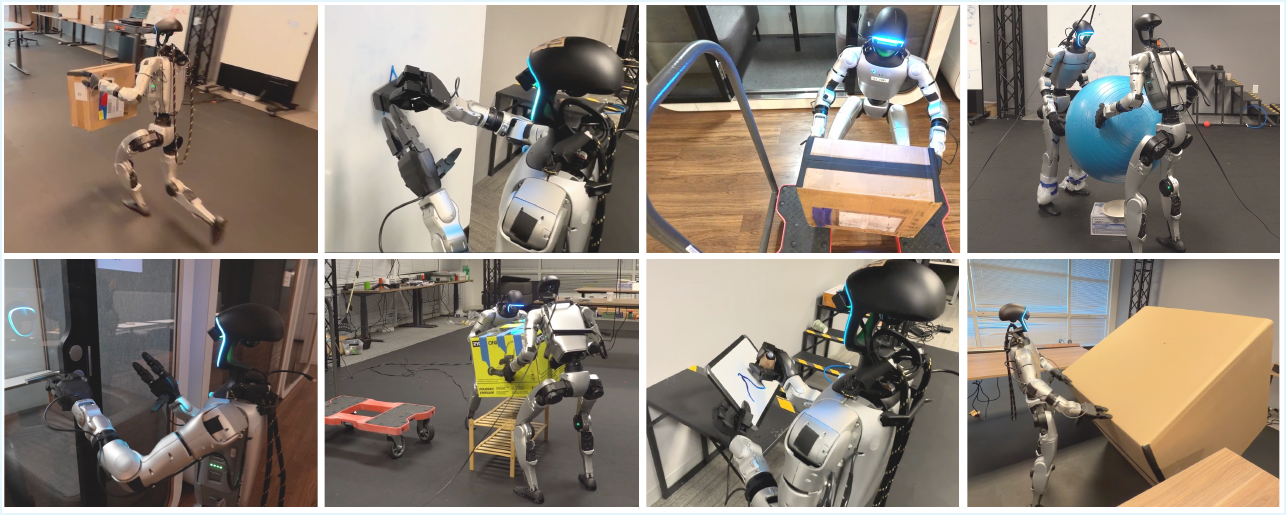}
  \captionof{figure}{CHIP enables humanoid robots to perform manipulation tasks that require force control, such as wiping a whiteboard, door opening, lifting boxes, flipping a large object, multi-robot collaboration, and writing. It also achieves both compliant and agile humanoid control, such as holding a box while running.}
  \label{fig:teaser}
  \vspace{-10pt}
\end{strip}

\begin{abstract}
Recent progress in humanoid robots has unlocked agile locomotion skills, including backflipping, running, and crawling. Yet it remains challenging for a humanoid robot to perform forceful manipulation tasks such as moving objects, wiping, and pushing a cart. We propose adaptive \underline{C}ompliance \underline{H}umanoid control through h\underline{I}sight \underline{P}erturbation (CHIP), a plug-and-play module that enables controllable end-effector stiffness while preserving agile tracking of dynamic reference motions. CHIP is easy to implement and requires neither data augmentation nor additional reward tuning. We show that a generalist motion-tracking controller trained with CHIP can perform a diverse set of forceful manipulation tasks that require different end-effector compliance, such as multi-robot collaboration, wiping, box delivery, and door opening.
\projecturl
\end{abstract}

\section{Introduction}
Recent advances in humanoid robots have enabled impressive progress in agile locomotion skills, allowing robots to walk, run, backflip, and even dance with human-like fluidity. However, despite these achievements, humanoid robots remain limited in manipulation and can often interact only with lightweight objects. A key challenge in humanoid manipulation is controlling robots to apply consistent and controllable interaction forces. In traditional tabletop manipulators, this problem is typically addressed using a model-based compliance force controller, such as impedance or admittance control. Yet, the agility of modern humanoids primarily stems from reinforcement learning (RL) based controllers, which lack a generalizable formulation for compliant force control.

RL-based compliance controllers have shown promising results on quadruped robots \cite{unifp, facet} by tracking motion spring-damper-like reference motion. However, these methods require substantial effort to generate large volumes of synthetic data that emulate spring-damper end-effector dynamics. Extending these approaches to humanoids is challenging, as it is difficult to ensure that such synthetic perturbation data remains within the distribution of natural human motion. In parallel, advances in RL-based motion tracking \cite{beyondmimic, twist2, any2track} have demonstrated that learning from human demonstrations is a scalable approach for constructing general-purpose humanoid controllers. However, these methods do not provide a principled way to integrate compliant or impedance control into humanoid tracking frameworks. In this work, we pose the central question: \textbf{Can we develop a scalable and generalizable approach for reconciling the high-gain stiffness required for humanoid agile motions with the variable compliance needed for safe contact-rich manipulation?} Some concurrent efforts~\cite{gentlehumanoid, softmimic} attempt to extend FACET~\cite{facet} to humanoids by augmenting retargeted human motion data with synthetic observations per responses under force perturbations, and then relying on tracking rewards to encourage imitation of the augmented trajectories. Yet, as is common in RL-based motion tracking \cite{hover, omnih2o, twist2, amo, head}, reward computation typically depends on rich reference information such as link poses, velocities, joint positions, and joint velocities, which are challenging to modify reliably during data augmentation. 

Our key insight for tackling this problem is that reference motion can be interpreted, in hindsight, as the robot’s compliant response to perturbations, allowing it to be used directly for dense reward computation without modification. Rather than altering the reference trajectory, we modify the sparse tracking goals as input obsesration to the policy, which are far easier to edit in a controlled manner. Specifically, we subtract the effect of a perturbation from the original reference motion to define the tracking goal in hindsight, while keeping the reference motion intact to provide high-quality, dense tracking rewards.

To this end, we introduce \textbf{C}ompliant \textbf{H}umanoid natural control through h\textbf{I}ndsight \textbf{P}erturbation (CHIP). This method integrates reinforcement learning with adaptive compliance control and can be directly incorporated into existing humanoid motion tracking frameworks with minimal modification. We demonstrate that generalist humanoid motion trackers trained with CHIP can perform a diverse range of forceful manipulation tasks—such as wiping, writing, cart pushing, and door opening while preserving their agility in tasks like dancing, running, and squatting. Furthermore, for tasks that require world-space coordination among multiple robots, we train a global 3-point tracking policy (i.e., tracking the head and two hands) using CHIP, enabling compliant multi-robot collaboration, including synchronized grasps and cooperative object transport.
In summary, our core contributions are:
\begin{itemize}
    \item CHIP, a generalizable and scalable plug-and-play module that enables adaptive compliance control in a humanoid tracking framework through hindsight perturbation.
    \item A compliant and natural local 3-point tracking controller that unlocks forceful whole-body teleoperation and autonomous Vision Language Action (VLA) policy learning, while maintaining the agile motion capabilities.
    \item A compliant global 3-point tracking controller that unlocks stable multi-robot object grasping and moving.
\end{itemize}
\section{Related work}
\subsection{Variable and adaptive compliance manipulation}
Adaptive compliance control has been used in manipulation to handle contact uncertainty and solve tasks that require force control. Prior work \cite{vices, acp} shows that variable impedance control could enable a robot to perform tasks that require significant contact force control, such as door opening, or to maintain surface contact, such as wiping. Under contact uncertainty, an adaptive compliance controller can also help robots perform stable grasps \cite{springgrasp, hmc}. Most prior adaptive compliance control studies use a tabletop arm with a model-based impedance controller. In our work, we draw inspiration from task design in these works, but apply it to a learned adaptive compliance controller on a humanoid robot. 

\subsection{Compliant control of legged robot}
Recent success in legged locomotion and loco-manipulation benefits from large-scale reinforcement learning in simulations. RL motion tracking has become a pivotal task enabling general humanoid control. Policies trained purely for motion tracking, often inspired by frameworks such as DeepMimic~\cite{deepmimic}, tend to be stiff and fragile under force perturbations. When the objective treats any deviation from a reference motion as an error that must be aggressively corrected, the robot generates large, uncontrolled forces during unexpected contact. To address this, recent research has pursued two distinct paradigms: learning to resist external forces or learning to comply with them.

\textbf{Learning to Resist Disturbances}. One line of work develops controllers designed to maintain tracking performance despite large, unknown external forces. These methods typically train policies with randomized external pushes to encourage robustness. Approaches that combine motion-tracking rewards with random external forces instruct the robot to adhere to the reference trajectory despite interaction forces \cite{ze2025twist, fu2022dwbc, zhang2025falcon}. A prominent example is FALCON~\cite{zhang2025falcon}, which uses a torque-aware 3D force curriculum, enabling the robot to gradually learn in simulation to resist strong disturbances and pull heavy carts. While effective for applications requiring forceful opposition, this stiffness-centric approach is ill-suited for contact-rich manipulation scenarios—such as robot-human collaboration, wiping, or handling fragile objects—which require the robot to yield in a controlled, spring-like manner.

\textbf{Learning to Comply with Interactions}. In contrast to force rejection, a second line of research aims to learn controllers that yield to external forces. One approach creates impedance behaviors "from scratch" using RL. For example, UniFP~\cite{unifp} and FACET~\cite{facet} train unified policies to mimic target spring-mass-damper dynamics. UniFP learns a force estimator for admittance control, while FACET modulates policy stiffness as a control input, enabling tasks such as wiping using only proprioception. However, because they are primarily designed for quadruped robots with a single end-effector, scaling these methods to humanoids is challenging because generating synthetic compliance data that matches the distribution of natural human movement is difficult.

To sidestep the difficulty of discovering behaviors from scratch, prior work has attempted to build compliance into motion-tracking frameworks. SoftMimic~\cite{softmimic} proposes a learning-from-examples framework, using an offline inverse kinematics (IK) solver to generate an augmented dataset of stylistically compliant motions. An RL policy then learns to reproduce these pre-authored responses. Similarly, GentleHumanoid~\cite{gentlehumanoid} augments keypoint reference motions with desired spring-damper dynamics during reward calculation. However, this creates an optimization conflict between motion tracking rewards and compliance objectives, often leading to degraded tracking accuracy or reduced agility. To make things worse, relying on offline data augmentation or reward tuning limits their applicability for scaling up to large, diverse motion datasets.

Our work builds upon these insights but proposes a more direct and scalable alternative. We introduce CHIP, a training recipe that integrates principled impedance-control objectives directly into the online RL loop via \textit{hindsight perturbation}, without compromising tracking performance. CHIP obviates the need for the extensive synthetic interaction curriculum \cite{unifp, facet} or the offline motion augmentation pipelines of SoftMimic~\cite{softmimic} and GentleHumanoid~\cite{gentlehumanoid}. By specifically targeting modifications in the input space, our method offers a simple yet powerful way to equip standard humanoid tracking systems with adaptive compliance, narrowing the gap between agile motion and safe, controllable physical interaction.

\subsection{Humanoid control interfaces}
Recent advances in reinforcement learning for humanoid whole-body control have produced remarkable results in agile locomotion~\cite{beyondmimic, kungfubot, hub, sonic} and whole-body manipulation~\cite{ze2025twist, amo, homie}. Extending motion-tracking approaches, several methods~\cite{ze2025twist, gmt} use human whole-body motion—encompassing both link- and joint-level states—to drive humanoid robot control. However, such control interfaces are complicated to use for both teleoperation and specifying high-level task commands. To simplify humanoid control, several methods \cite{homie, amo} adopt a decoupled upper-lower-body policy, in which the robot receives root-velocity and height commands for locomotion and upper-body target-motion commands for manipulation. Other approaches explore keypoint-based control \cite{omnih2o, hover, sonic, head, twist2, clone}, enabling intuitive teleoperation by mapping head and wrist movements from a VR headset to humanoid motion. CHIP upgrades humanoid keypoint control with an adaptive-compliance module. It is compatible with both local keypoint tracking, as in SONIC~\cite{sonic}, OmniH2O~\cite{omnih2o}, and global keypoint tracking \cite{clone, head}, while additionally providing a plug-and-play module that directly integrates task-aware control into existing keypoint-based frameworks.

\section{Method}

\begin{figure*}[t!]
  \centering
  \includegraphics[width=0.9\linewidth]{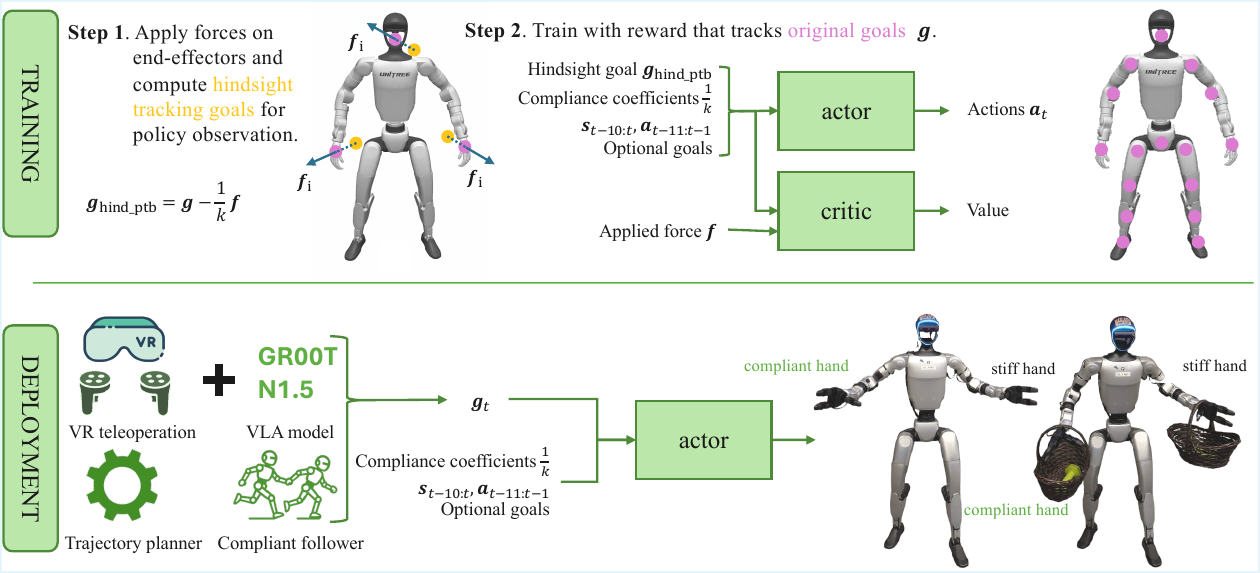}
  \caption{
  Overview of training and deployment of CHIP: this general formulation takes tracking targets, end-effector compliance coefficients, and state history as default inputs. For the local tracking policy, we provide an additional lower-body reference pose, generated via the kinematic planner used in SONIC~\cite{sonic}. It adapts end-effector compliance based on the input compliance coefficient and force implicitly estimated from proprioception history. During training, the policy observes recovered keypoint poses by removing the hindsight perturbation, and tracking rewards are calculated from the original reference motion. The trained adaptive compliance policy could be controlled via multiple interfaces, such as VR teleoperation, a kinematic planner, and a vision-language-action (VLA) model.
  }
  \label{fig:model}
\end{figure*}

\begin{figure}[h!]
  \centering
  \includegraphics[width=\linewidth]{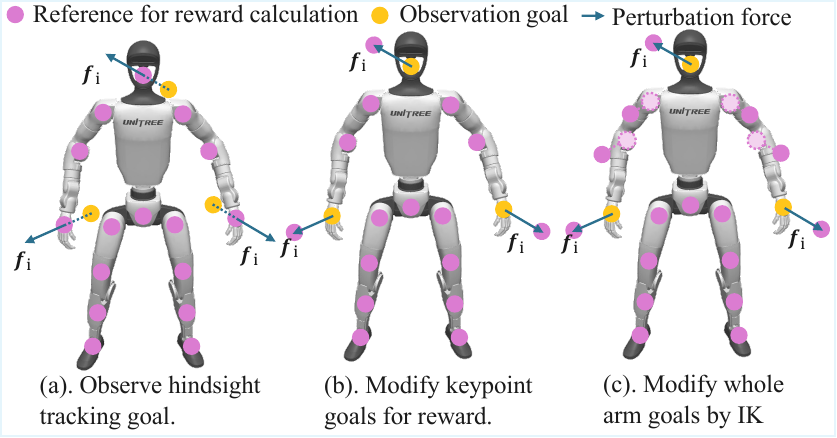}
  \caption{
  Different formulations for adaptive compliant tracking policy training. 
  (a) Ours (CHIP), the policy observes hindsight tracking goal, a modified end-effectors goal in the opposite direction of the perturbation force. 
  (b) Modify keypoint goals for reward as in GentleHumanoid~\cite{gentlehumanoid}. 
  (c) Edit entire kinematic reference motion to account for end-effector perturbation as in SoftMimic~\cite{softmimic}.
  }
  \label{fig:hindsight}
  \vspace{-15pt}
\end{figure}

% Need to add proprioception info. 
Our method is a lightweight plug-and-play module that integrates seamlessly into any keypoint-based motion-tracking framework and enables adaptive whole-body control policies with end-effector-level compliance. As shown in Fig.~\ref{fig:model}, CHIP takes the 3-point tracking goal $\bm{g}_t$, end-effector compliance coefficients $\frac{1}{k}$, proprioception $\bm{s}_{t-10:t}$, and past actions $\bm{a}_{t-11:t-1}$ as input, and outputs actions that enables a humanoid robot to track the 3-point targets while yielding to external perturbations, with the desired level of compliance modulated by the continuous input coefficients.

%The policy receives tracking goals \(\bm{g}\), which specify desired keypoint poses, and relies on a collection of dense reward terms to drive the robot's joints and links to align with those in the reference motion (Appendix~\ref{appd:rewards}).

The core of CHIP lies in modifying the training procedure within the motion tracking framework. During training, we apply a perturbation force $\bm{f}_i$ in a random direction to the end-effector $i$ for a random period of time. The policy also receives compliance coefficients $\frac{1}{k}$ that allow it to follow an impedance control law, mimicking spring--damper dynamics under external disturbances.

Prior work \cite{unifp, facet} and recent concurrent work \cite{gentlehumanoid, softmimic} train policies $\pi(\bm{s}, \bm{g})$ that observe tracking goals $\bm{g}$ from the original reference motion, but use tracking goals modified by perturbations, $\bm{g}_\text{ptb}$, to calculate the reward that encourages the robot to track the desired compliant response under force perturbations rather than the original motions (Fig.~\ref{fig:hindsight}(b)):
\[
r(\bm{s}, \bm{g}_\text{ptb}) = \exp(-||\bm{x}_\text{eef} - \bm{g}_\text{ptb}||), \qquad \bm{g}_\text{ptb} = \bm{g} + \frac{1}{k}\bm{f},
\]
where $\bm{x}_\text{eef}$ is the current end-effector pose. This approach requires modifying the reference trajectory while maintaining its feasibility (Fig.~\ref{fig:hindsight}(c)), a process that becomes increasingly difficult for dynamic skills such as running, where motion editing needs to satisfy both kinematic and dynamic constraints.

Instead, our method modifies the \emph{observed tracking goals} as input to the policy in a hindsight manner, 
\[
\pi(\bm{s}, \bm{g}_\text{hind\_ptb}), \qquad
\bm{g}_\text{hind\_ptb} = \bm{g} - \frac{1}{k}\bm{f},
\]
such that the original goals are reinterpreted as the perturbation's resulting motion and are directly used to compute the reward function, $r(\bm{s},\bm{g})$ (Fig.~\ref{fig:hindsight}(a)), eliminating the need to alter the reference motion.

This design has several advantages:
\begin{itemize}
    \item It takes less effort to modify the sparse goal in the observation than the dense reference motion in reward computation, allowing such modification to be done online during training.
    \item Robots are always encouraged to produce motion in the distribution of reference motion, which improves motion naturalness.
    \item Robots are exposed to observations from both the motion dataset and outside it, thereby improving policy robustness at test time.
\end{itemize}

This formulation streamlines training while enhancing robustness and agility. Unlike methods such as~\cite{unifp, facet}, which explicitly learn an external force estimator, our approach uses a single policy model optimized solely with the PPO objective. The policy implicitly acquires force awareness through proprioceptive observations and past actions. To further improve sensitivity to external perturbations during training, we provide the critic with the ground-truth external force as a privileged observation. In addition, we supply both the actor and critic with a 10-step history of proprioception and past actions, allowing the policy to infer perturbation-related information from noisy observations. At test time, the policy only receives the tracking goal $\bm{g}$ and proprioception. It implicitly estimates the perturbation force and generates actions that counteract disturbances according to the commanded compliance. To enable compliant following, similar to FACET~\cite{facet}, we also implement a damper model on the target end-effector pose. Under an applied force, the end-effector pose $\bm{x}_{\text{eef}}$ deviates from the previous target $\bm{g}_{t-1}$; we update the target using:
\[
\bm{g}_{t} = \alpha \bm{x}_{\text{eef}} + (1-\alpha) \bm{g}_{t-1}.
\]
This update allows the robot to adjust its target in response to force-induced deviations smoothly.

\section{Applications}
We showcase two applications that utilize our compliance humanoid control: multi-humanoid manipulation and compliant teleoperation. The former application requires 3-point tracking in the global frame, and the latter depends only on a local tracking policy.

\subsection{Compliant multi-robot grasping}

For tasks requiring global coordination, such as multi-robot collaborative grasping or object transport, we train a whole-body control policy that tracks the 6D pose of the head and the 3D positions of both wrists in the world frame:
\[ \bm{g} = 
\{\bm{p}_\text{head}^\text{ref}, \bm{r}_\text{head}^\text{ref}, \bm{p}_\text{lwrist}^\text{ref}, \bm{p}_\text{rwrist}^\text{ref}\}.
\]
This target space provides a simple interface for commanding robot end-effectors to reach desired locations in the global coordinate frame. Training uses a global tracking reward that penalizes the 3-point deviation between the robot and the reference motion in the world frame. However, because 3-point global poses provide only sparse information, it remains ambiguous for the robot to determine whether to walk or extend its arm and head when the 3-point target moves. We use 0.2 seconds of future information (5 frames, four frame skip per step) to resolve such ambiguity.

With an adaptive, compliant global 3-point controller, multiple robots can position their end-effectors in the world frame and apply desired forces, enabling multi-robot collaborative grasping and the movement of large objects beyond the grasping capability of a single robot.

To obtain a two-robot grasp, we extend the dexterous compliant grasp generation algorithm SpringGrasp~\cite{springgrasp} to a multi-humanoid collaborative grasp setting. We solve an optimization problem to obtain the 6D head poses and 3D wrist positions of two robots:
\[
\{\bm{p}^i_\text{lwrist}, \bm{p}^i_\text{rwrist}, \bm{p}^i_\text{head}, \bm{r}^i_\text{head}\}, \qquad i \in \{1,2\},
\]
ensuring a stable, collision-free, compliant grasp.

We then plan a minimum-curvature trajectory between the current and target poses, interpolating wrist targets along the path. The grasping procedure consists of three phases:
\begin{enumerate}
    \item \textbf{Approach}: robots follow the trajectory to pre-grasp poses near the object.
    \item \textbf{Grasp}: wrist targets penetrate slightly into the object to establish a compliant grasp.
    \item \textbf{Lift}: head and wrist targets move upward to lift the object collaboratively.
\end{enumerate}
After the object has been lifted, we can simultaneously translate the 3-point target poses of both robots to move the grasped large object. For multi-robot collaboration to work in diverse environments, we require high-frequency, infrastructure-free robot localization. Following BeyondMimic~\cite{beyondmimic}, we combine:
\begin{itemize}
    \item Foot-odometer based root velocity estimation at 50 Hz,
    \item Fast-LIO2 based pose updates at 10 Hz~\cite{fastlio2},
    \item Root-IMU angular velocity at 50 Hz,
\end{itemize}
yielding a fused 50\,Hz root state estimate.

Both robots also need to share a consistent global frame. We first built an environment point map using Fast-LIO2. During initialization, each robot registers its live point cloud to the prebuilt map, aligning its coordinate frames.

\subsection{Teleoperation and vision language action model training}
\subsubsection{Local 3-point tracking policy}

For tasks that do not require global coordination in the task space, we train a local policy that tracks the SE(3) poses of the head and wrists relative to the robot's root frame, while also tracking lower-body joint positions and velocities to prevent drift in the absence of global information. Such a formulation acts as a decoupled policy that separates upper-body movement from lower-body locomotion at the command level. 
% \karen{One sentence to explain why the lower body joint is helpful in the absence of global information.}. 
\[ \bm{g} = 
\{\bm{p}_\text{head}^\text{ref}, \bm{r}_\text{head}^\text{ref}, \bm{p}_\text{lwrist}^\text{ref}, \bm{r}_\text{lwrist}^\text{ref}, \bm{p}_\text{rwrist}^\text{ref}, \bm{r}_\text{rwrist}^\text{ref},\bm{q}_\text{lower}^\text{ref}\}.
\]
The reward minimizes the difference between the reference and robot 3-point poses expressed in the root frame. At deployment time, desired 3-point poses are provided either through a VR-based teleoperation system or via rollouts from a Vision-Language-Action (VLA) policy. Lower-body trajectories are generated by a kinematic planner that receives commands for root velocity, heading, and height.

\begin{figure}[h!]
  \centering
  \includegraphics[width=0.9\linewidth]{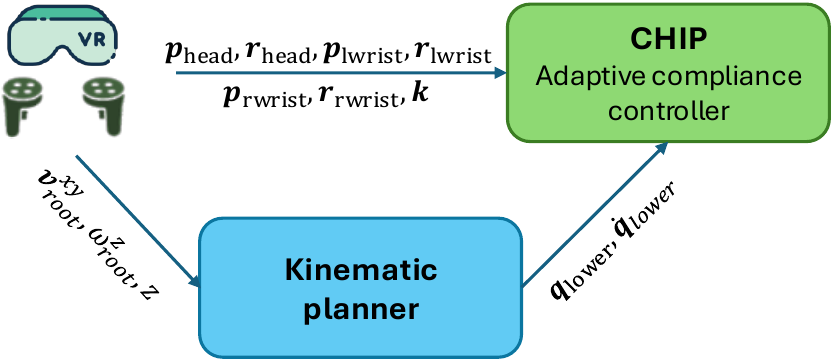}
  \caption{Teleoperation interface, VR directly gives wrist and head pose commands to the whole body controller, root commands are sent to the kinematic planner to generate lower body joint motion.}
  \label{fig:teleop}
\end{figure}
For tasks that do not require global localization, the compliant local 3-point policy is straightforward to deploy. Following SONIC~\cite{sonic}, the robot is teleoperated using VR-tracked wrist poses along with root velocity and height commands from VR joysticks as shown in Fig.~\ref{fig:teleop}. Lower-body control is handled by a kinematic planner driven by these commands. The kinematic planner takes root velocity, angular velocity, and height as commands and outputs planned joint positions and velocities. Compliance coefficients for each end-effector are adjusted by the operator in VR. Teleoperation sessions collect paired action data and compliance coefficients from the teleoperation interface, as well as binocular RGB observations from the robot's egocentric OAK camera. These data are used to finetune the GR00T N1.5~\cite{groot} VLA model. 
\section{Experiments and Results}
\begin{figure*}[t]
  \centering
  \includegraphics[width=0.75\linewidth]{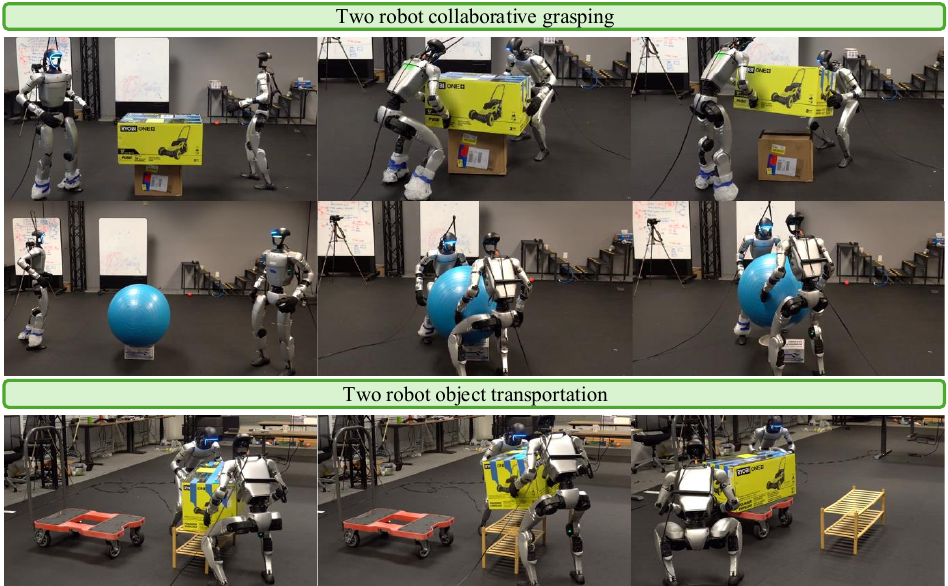}
  \caption{\textbf{Multi-robot grasp and object transportation:} In the grasping experiment, two robots walk toward the object, grasp and lift the object with their end-effectors; Two robots can also move the grasped object and place it in a new location.}
  \label{fig:mrgrasp}
  \vspace{-10pt}
\end{figure*}
We conducted an extensive experiment to answer the following research questions.
\begin{itemize}
\item Q1: How well does CHIP track position and force targets?
\item Q2: What task(s) do adaptive compliance unlock through teleoperation?
\item Q3: How does CHIP facilitate multi-robot collaboration?
\item Q4: Can CHIP enable VLA models to perform autonomous forceful manipulation?
\item \added{Q5: What is the advantage of the continuous compliance command space used in CHIP?}
\end{itemize}

\subsection{Experiment setup}
We used the Unitree G1 as our test platform, and an OAK-D W camera was mounted on the robot's head to provide egocentric binocular vision. All policies were trained for 4 days on 64 Nvidia L40S GPUs, with 4096 environments per GPU. During deployment, the local 3-point tracking policy, kinematic motion planner, and teleoperation server were running on the Unitree G1 Jetson NX onboard computer with TensorRT acceleration. The global 3-point tracking policy, together with the state estimator, was running on a desktop machine with an i7-13700K CPU and an RTX 3090 GPU. We use the same desktop computer to run VLA policies.
\subsection{Tracking performance of CHIP}

\begin{table}[t]
\centering
\footnotesize
\renewcommand{\arraystretch}{1.10}
\setlength{\tabcolsep}{4pt}

\begin{tabular}{>{\arraybackslash}p{0.28\columnwidth}cccc}
\toprule
\textbf{Method / Error} &
\textbf{$\epsilon_p^g$ [m]} &
\textbf{$\epsilon_r^g$ [rad]} &
\textbf{$\epsilon_p^l$ [m]} &
\textbf{$\epsilon_r^l$ [rad]} \\
\midrule
\makecell[l]{Ours\\{\scriptsize($1/k=0.00$)}} &
0.08 (0.04) & 0.09 (0.04) & 0.02 (0.02) & 0.09 (0.06) \\
\makecell[l]{Ours\\{\scriptsize($1/k=0.02$)}} &
0.08 (0.05) & 0.08 (0.05) & 0.02 (0.02) & 0.10 (0.06) \\
\makecell[l]{Ours\\{\scriptsize($1/k=0.05$)}} &
0.08 (0.04) & 0.08 (0.04) & 0.02 (0.02) & 0.11 (0.06) \\
\makecell[l]{Gentle\-humanoid\\{\scriptsize[stiff: max force=15N]}} &
- & - & 0.04 (0.03) & 0.15 (0.08) \\
\makecell[l]{Gentle\-humanoid\\{\scriptsize[compliant: max force=5N]}} &
- & - & 0.05 (0.04) & 0.16 (0.10) \\
FALCON &
0.09 (0.04) & 0.08 (0.04) & 0.02 (0.02) & 0.08 (0.05) \\
No perturb force &
0.11 (0.04) & 0.09 (0.05) & 0.02 (0.02) & 0.08 (0.05) \\
\bottomrule
\end{tabular}

\caption{Tracking errors are measured in the MuJoCo simulator without external force perturbation. The \textit{No perturb force} baseline does not train with force perturbation at all; the FALCON~\cite{zhang2025falcon} is trained with force perturbations on the end-effector but does not observe compliance. \added{Gentlehumanoid~\cite{gentlehumanoid} observes compliance commands but uses reward-shaping to get the compliant behavior.} Entries are reported as mean (standard deviation). Metrics: For global policy $\epsilon_p^g$ -- global position error [m], $\epsilon_r^g$ -- global orientation error [rad]; for local policy $\epsilon_p^l$ -- local position error [m], and $\epsilon_r^l$ -- local orientation error [rad].}
\label{tab:accuracy}
\vspace{-15pt}
\end{table}

We analyze the tracking accuracy and the robot's response to force perturbations under different end-effector compliance settings. Tab.~\ref{tab:accuracy} shows the tracking accuracy measured over 100 trajectories sampled from the TWIST dataset~\cite{ze2025twist}; it shows that the use of hindsight perturbation training yields position tracking on par with the no-compliance \added{(FALCON~\cite{zhang2025falcon})} and no-force-perturbation baseline, with only a slight impact on orientation tracking performance. \deleted{Additionally, our method consistently outperforms the compliant baseline (GentleHumanoid~\cite{gentlehumanoid}) across different compliance levels at evaluation time.} \replaced{Meanwhile, Tab.~\ref{tab:accuracy} shows}{Meanwhile, the results show} that varying end-effector stiffness does not significantly affect the tracking performance of our method, indicating that the policy can distinguish between \replaced{the internal joint's driving force}{the joint's internal driving force} required for agile performance and external perturbation forces. \added{Compared with another compliant baseline, Gentlehumanoid~\cite{gentlehumanoid}, which removes dense joint-level tracking reward to achieve compliance behavior, CHIP outperforms it in terms of higher tracking accuracy.} \added{This result clearly shows that the design choice in CHIP, which modifies observations via hindsight perturbation rather than reward tuning, is a better approach to achieve both compliance behavior and accurate general tracking performance.}

\subsection{Multi-robot grasping and object transportation}

\begin{figure}[h]
\vspace{-10pt}
  \centering
  \includegraphics[width=0.95\linewidth]{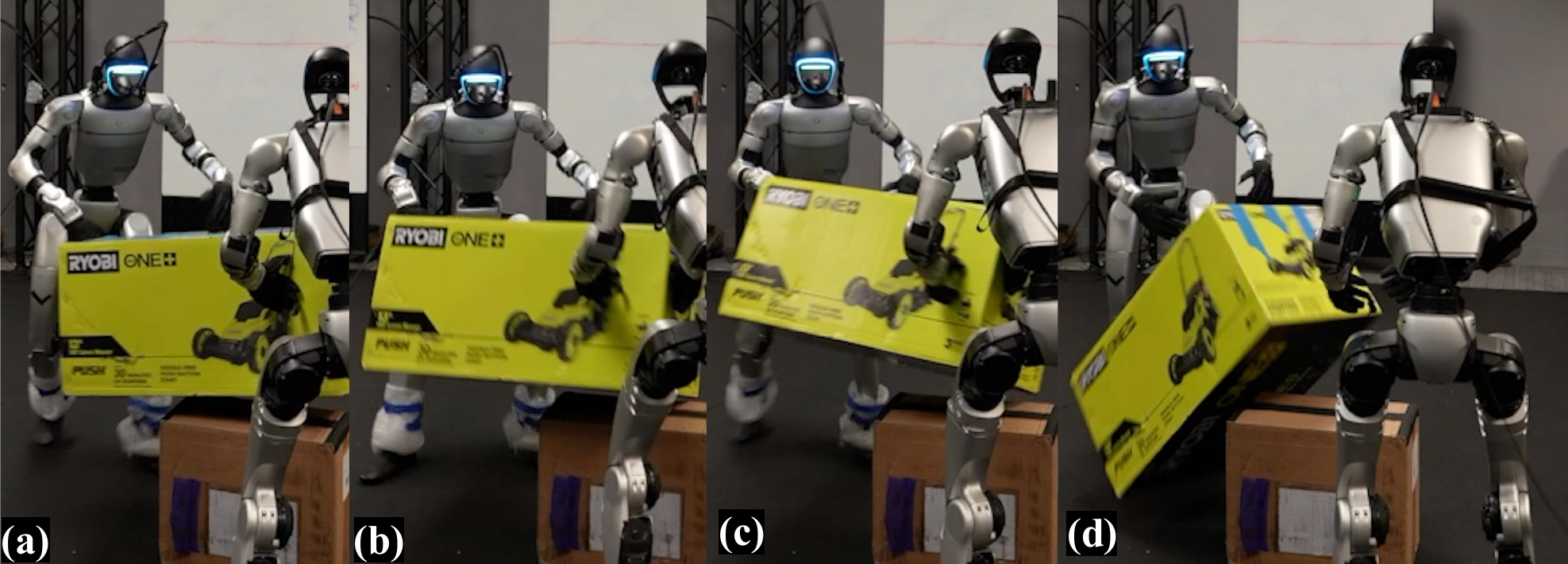}
  \caption{
      \added{
      \textbf{Multi-robot grasp failure cases} of box lifting caused by an unstable contact with a non-compliant controller (FALCON~\cite{zhang2025falcon}).
      }
  }
  \label{fig:multi_robot_failure}
  \vspace{-10pt}
\end{figure}

To demonstrate the effectiveness of our adaptive compliance controller for multi-robot collaborative grasp, we evaluated our method on the collaborative grasping of two objects of different heights, as shown in Fig.~\ref{fig:mrgrasp}. Robots first approached and grasped the object with their end-effectors, and the grasp was considered successful if they could lift it. Tab.~\ref{tab:grasp} shows the grasping success rate, and Fig. ~\ref{fig:multi_robot_failure} demonstrates how a non-compliant policy produces an unstable collaborative grasping. In general, using an adaptive compliance controller (tuned to 1/k=0.04 to balance the lifting force and the safety of the interaction) achieves an average grasp success rate of 80\%, which is +75\% higher than that of an always-stiff controller and +40\% higher than that of a controller without force perturbation. In failure cases with our method, the main reason was the robot's knee colliding with the object before grasping, which did not fully showcase our controller's capabilities. We noticed that an always-stiff controller performed worse than the controller trained without force, because a stiff controller usually applied a much larger force to push the object, creating a large opposing force between the two robots and the object, causing the object to destabilize immediately after the contact. We also show that after grasping the object, robots can stably move the object and place it in a new location, as shown in Fig.~\ref{fig:mrgrasp}. We used the keyboard command to translate end-effector target locations to move the object.
\begin{table}[!ht]
    \centering
    \resizebox{\linewidth}{!}{
    \begin{tabular}{l|ccc}
    \toprule
        Setting / Method  & Ours & FALCON(No compliance) & No force perturbation \\ \hline
        Box - 18 cm     & 0.6 & 0.0 & 0.4 \\
        Box - 35 cm     & 0.8 & 0.0 & 0.4 \\
        Sphere - 12 cm  & 1.0 & 0.2 & 0.4 \\
        Sphere - 35 cm  & 0.8 & 0.0 & 0.4 \\ 
        Average & 0.8 & 0.05 & 0.4 \\
    \bottomrule
    \end{tabular}}
    \caption{Multi-robot grasp success rate; all settings were evaluated over five experiments.}
    \label{tab:grasp}
    \vspace{-10pt}
\end{table}

\begin{figure*}[h!]
  \centering
  \includegraphics[width=0.95\linewidth]{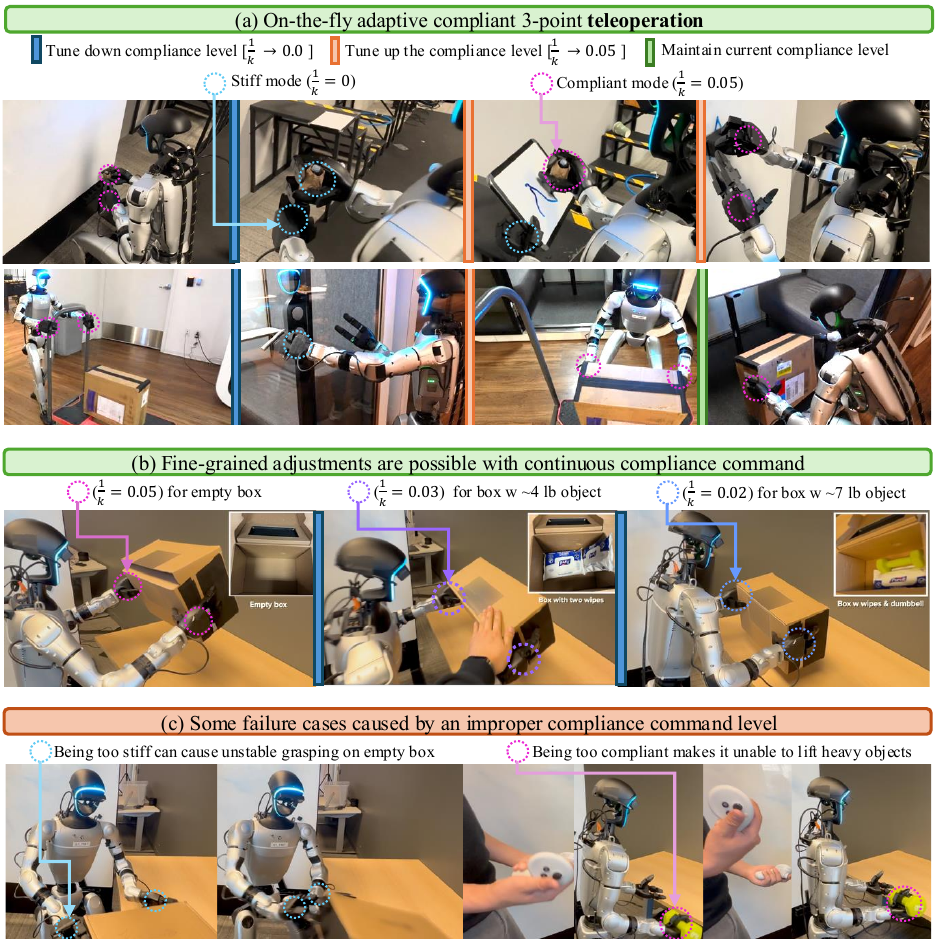}
  \caption{\textbf{(a) On-the-fly adaptive compliant 3-point teleoperation.} Hand compliance is annotated by $\frac{1}{k}$. Dashed cyan and magenta rings denote stiff mode ($\frac{1}{k}=0$) and compliant mode ($\frac{1}{k}=0.05$), respectively. Vertical bars indicate online compliance tuning commands (blue: tune down $\frac{1}{k}\!\rightarrow\!0$; orange: tune up $\frac{1}{k}\!\rightarrow\!0.05$; green: maintain). The unloading task benefits from compliant behavior during cart pushing and box grasping, but stiff behavior during door opening; the writing \& wiping task alternates between stiff pen opening and bimanual hybrid compliance. \textbf{(b) Fine-grained compliance adjustments} to make the controller \textit{stiff-just-at-needed-level}. When the box gets heavy, we can on-the-fly decrease $\frac{1}{k}$ from 0.05 to 0.02 to ensure the robot is stiff enough to lift the object while remaining stable even under external perturbation shown in (b)-middle frame. \added{\textbf{(c) Improper compliance command level} can cause clear failure. For empty-box lifting, using stiff mode results in unstable grasping even in teleop mode. However, we also need a stiff mode to generate sufficient force to lift heavy objects.}
  }
  \vspace{-15pt}
  \label{fig:teleop}
\end{figure*}

\subsection{Teleoperation} \label{hardware}
We demonstrated that the compliant local 3-point tracking controller can be teleoperated to accomplish a wide variety of contact-rich manipulation tasks, as illustrated in Fig.~\ref{fig:teleop}. The same controller supports both compliance-demanding tasks—such as lifting, transporting boxes, and wiping surfaces- as well as force-demanding tasks—such as opening a spring-loaded door, actuating a gantry, or flipping heavy boxes. 

We further show that it can complete tasks requiring different compliance settings on both hands, for example, holding a small whiteboard rigidly with one hand ($\frac{1}{k}=0.001$) while writing on it with the other ($\frac{1}{k}=0.05$). The user can also adjust compliance online during long-horizon tasks, such as first opening a marker with a stiff end-effector and then writing on the whiteboard in a compliant manner. Additional results are shown in our video and on our website.

\begin{figure}[!h]
  \centering
  \includegraphics[width=\linewidth]{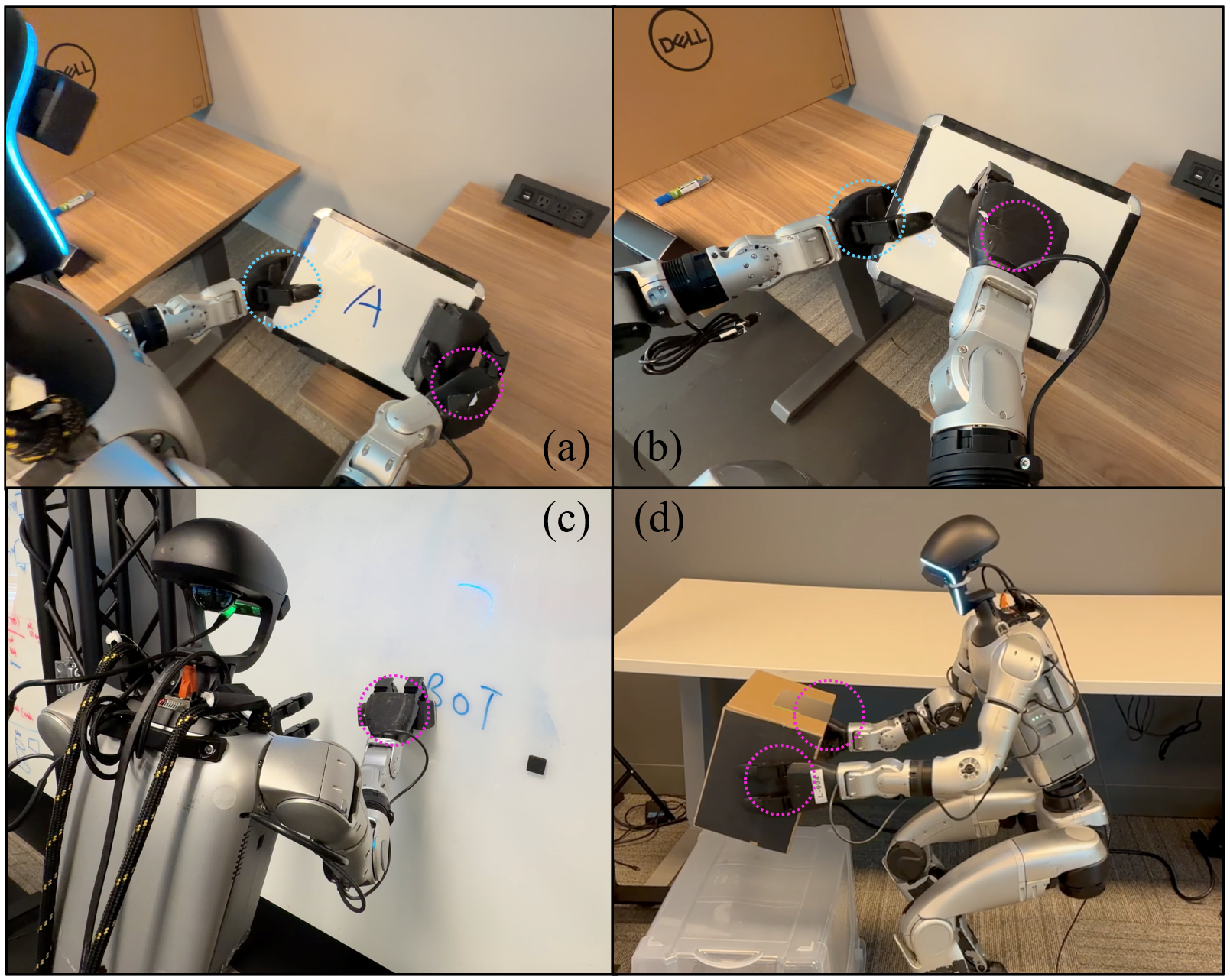}
  \caption{\textbf{Autonomous VLA Tasks.} We fine-tune VLA policy on three contact-rich tasks requiring different end-effector stiffness: (c) \textbf{Single-Arm Wiping}, requiring the robot to maintain contact with surface; (a,b) \textbf{Holding and Wiping}, robot must rigidly hold a small whiteboard with one hand while the other hand wipes compliantly; and (d) \textbf{Bimanual Pick and Place}, utilizing dual-arm compliance to ensure stability while grasping and moving a box.}
%   The purple dashed circle represents the compliance
% coefficient of 0.05, while the blue dashed circle represents the compliance
% coefficient of 0.}
  \label{fig:vla}
  \vspace{-10pt}
\end{figure}

\subsection{Autonomous VLA} \label{sec:autonomous_vla}
% Using data collected from teleoperation, we fine-tuned a VLA policy to perform tasks autonomously. For the large whiteboard wiping task, we collected 400 teleoperated trajectories. During data collection, the wiping hand compliance coefficient was set to 0.05. After fine-tuning, the robot was able to complete the task autonomously, as shown in Fig.~\ref{fig:teleop}. We evaluated the policy over 10 autonomous rollouts. A rollout was considered successful if the robot erased all text on the whiteboard within 2 minutes. Under this criterion, the policy achieved a 60\% success rate. The dominant failure modes were (i) the text being occluded by the robot hand and (ii) the text leaving the field of view of the robot-mounted camera.

% We also demonstrate that the VLA model can exploit different stiffness settings for different end-effectors. As illustrated in Fig.~\ref{fig:teleop}, the hand holding the small whiteboard was commanded with high stiffness, while the wiping hand used lower stiffness to maintain compliant contact. Using 200 teleoperated trajectories for this bimanual setup, we fine-tuned a separate VLA policy that enables the robot to autonomously hold and wipe the whiteboard, achieving an 80\% success rate.

\added{
We further show that, by using teleoperation to collect manipulation data with appropriate compliance, we can enable VLA policy to accomplish some contact-rich manipulation tasks autonomously. There is a total of three tasks chosen to demonstrate contact-rich manipulation skills in Fig.~\ref {fig:vla}: (i) a single-arm wiping skill requiring compliance in one end-effector, (ii) an holding and wiping setting where different end-effectors require different stiffness levels, and (iii) a bimanual pick and place task where compliance in both hands is desired for grasp stability. For each task, we evaluate the success rate across 20 policy rollouts; a rollout is considered a failure if the robot does not finish the task within 2 minutes.
}
% Using teleoperated demonstrations, we fine-tuned vision-language-action (VLA) policies to perform autonomously 
% \karen{Do we have comparisons with baselines for the following exps? -- Ziang: We will enhance the Tab~\ref{tab:grasp} related section to prove that we need compliance in some cases.}

\added{
\textbf{Wiping of the large whiteboard}. The robot walks to the front of the whiteboard, grasps a wiper with the right hand, and erases text while maintaining compliant contact. We collected 400 teleoperated trajectories with a wiping-hand compliance coefficient of 0.05, then fine-tuned a VLA policy. A rollout is considered successful if the robot removes all text within the specified time. According to this criterion, the policy achieved a 60\% success rate. Even in some cases, a robot cannot successfully wipe out all text, but it can maintain stable contact during wiping. The dominant failure modes were (i) text that was occluded by the robot hand and (ii) remaining text that had left the field of view of the robot-mounted camera.
}

\added{
\textbf{Bimanual small whiteboard wiping}. We then increase task complexity by requiring end-effector-specific stiffness within a single bimanual skill. As illustrated in Fig.~\ref{fig:vla}, the robot holds the small whiteboard with its left hand at high stiffness, while making the right wiping hand use lower stiffness to maintain compliant contact. Using 200 teleoperated trajectories, we fine-tuned a separate VLA policy and achieved an 80\% success rate.
}

\added{
\textbf{Box lifting}. Finally, we consider a task where bimanual compliance is clearly beneficial. The robot squats to retrieve a box from a low position (Fig.~\ref{fig:vla}), lifts it, turns 90°, and places it on a table. We use lower stiffness in both hands to improve grasp stability during robot motion. We collected 200 teleoperated trajectories and fine-tuned a separate VLA policy. The policy achieved a 90\% success rate; the main failure mode was an unsuccessful initial grasp.
}

\subsection{Analysis: The Necessity of Continuous Compliance}
\added{
We observe a critical trade-off in the box-lifting task (Fig.~\ref{fig:teleop}-b): while high compliance enhances contact stability, heavier payloads require greater stiffness to maintain force transmissibility. CHIP's continuous spectrum resolves this via:
}
\begin{itemize}
    \item \added{\textbf{Optimized Force-Stability Trade-off:} The controller can be tuned to be strictly stiff enough to lift the load while remaining compliant enough to absorb shocks, avoiding the brittleness of always-stiff policies. Compared with the max-cutoff-force (5-15 N) design from Gentlehumanoid, we can still enable different compliance behaviors even when the external force falls outside Gentlehumanoid's threshold.}
    \item \added{\textbf{Semantic Stiffness Adaptation:} Our continuous compliance command links to physically meaningful Hooke's law. Therefore, the spectrum design enables teleoperators to continuously modulate stiffness based on visual mass estimation, facilitating easy adaptability.}
\end{itemize}
\section{Limitations}
\added{
Although CHIP can equip keypoint-based humanoid tracking systems with an adaptive end-effector compliance input, our current design is constrained by the assumptions underlying sensing and modeling. Due to limitations in the proprioceptive accuracy of the Unitree G1 platform, our real-robot implementation currently employs \emph{unidirectional} stiffness to specify end-effector positional compliance. Lacking accurate wrist force--torque sensing further limits precise and versatile admittance control and may hinder fine-grained contact tasks such as peg-in-hole insertion. In addition, selecting appropriate compliance coefficients remains interface- and task-dependent (e.g., teleoperation/VLA/planner), and may require manual tuning when tasks alternate between force-demanding and precision phases. Finally, while we demonstrate collaborative grasping and object transportation across multiple robots, the evaluated scenarios are limited, and some failures (e.g., knee collisions before grasp) do not fully isolate the controller’s interaction capabilities. 
}
\section{Conclusion}
We introduced CHIP, a framework for learning humanoid natural adaptive compliance through hindsight replay. CHIP serves as a lightweight plug-and-play module for keypoint-based humanoid motion-tracking systems, enabling end-effector-level adaptive compliance control with minimal modification to existing pipelines. We demonstrated that our method produces adaptive local compliance policies capable of executing force-demanding tasks such as wiping, opening the door, and carrying boxes. Moreover, CHIP supports a globally adaptive compliant 3-point tracking controller, enabling multiple humanoid robots to manipulate and transport large objects collaboratively.

\ifanonym
    \newpage
\else
    \section{Acknowledgement}
We thank Jeremy Chimienti, Tri Cao, Peter Pham, Justin Tran, Rajeev Varma, Smit Patel, Caleb Geballe, Runyu Ding, David Sami, Dennis Da, Tairan He, Zi Wang, Haoru Xue, and Wenli Xiao for their help and support during this project.
\fi

\bibliographystyle{unsrt}
\bibliography{biblio}
\clearpage

\subsection{Reward Design}
\label{appd:rewards}
Our tracking reward design is inspired by prior work \cite{sonic,beyondmimic}. Tab.~\ref{tab:rl_rewards} shows our reward terms and weights.

% OPTIMIZATION 1: Single column table with tighter vertical spacing
\begin{table}[h]
\centering
\setlength{\tabcolsep}{2pt} % Reduce column padding
\renewcommand{\arraystretch}{0.85} % Reduce row height
\resizebox{\columnwidth}{!}{ % Force fit to single column width
\begin{tabular}{l l c}
\toprule
\textbf{Reward term} & \textbf{Equation} & \textbf{Weight} \\
\midrule
\multicolumn{3}{l}{\textit{Tracking rewards $\mathcal{R}({\bs{s}^{\text{p}}_t}, \bs{s}^{\text{g}}_t)$}} \\
% TODO: please add a small gap between rows here -- smaller than \\
Root orientation & $r^{\text{root}}_{\text{ori}}(t)=\exp\!\big(- \|\bs{o}^p_{t,r}-\bs{o}^g_{t,r}\|_2^2 / 0.4^2\big)$ & 0.5 \\
Body link pos (rel.) & $r^{\text{body}}_{\text{pos}}(t)=\exp\!\Big(-\tfrac{1}{|\mathcal{B}|}\!\sum_{b\in\mathcal{B}}\|\bs{p}^{p,\text{rel}}_{t,b}-\bs{p}^{g,\text{rel}}_{t,b}\|_2^2 / 0.3^2\Big)$ & 1.0 \\
Body link ori (rel.) & $r^{\text{body}}_{\text{ori}}(t)=\exp\!\Big(-\tfrac{1}{|\mathcal{B}|}\!\sum_{b\in\mathcal{B}} \|\bs{o}^{p,\text{rel}}_{t,b}-\bs{o}^{g,\text{rel}}_{t,b}\|_2^2 / 0.4^2\Big)$ & 1.0 \\
Body link lin. vel & $r^{\text{body}}_{\text{lin}}(t)=\exp\!\Big(-\tfrac{1}{|\mathcal{B}|}\!\sum_{b\in\mathcal{B}}\|\bs{v}^p_{t,b}-\bs{v}^g_{t,b}\|_2^2 / 1.0^2\Big)$ & 1.0 \\
Body link ang. vel & $r^{\text{body}}_{\text{ang}}(t)=\exp\!\Big(-\tfrac{1}{|\mathcal{B}|}\!\sum_{b\in\mathcal{B}}\|\bs{\omega}^p_{t,b}-\bs{\omega}^g_{t,b}\|_2^2 / 3.14^2\Big)$ & 1.0 \\
3-point pos (rel.) & $r^{\text{3pt}}_{\text{pos}}(t)=\exp\!\Big(-\tfrac{1}{|\mathcal{B}|}\!\sum_{b\in\mathcal{B}}\|\bs{p}^{p,\text{rel}}_{t,b}-\bs{p}^{g,\text{rel}}_{t,b}\|_2^2 / 0.3^2\Big)$ & 1.0 \\
3-point ori (rel.) & $r^{\text{3pt}}_{\text{ori}}(t)=\exp\!\Big(-\tfrac{1}{|\mathcal{B}|}\!\sum_{b\in\mathcal{B}} \|\bs{o}^{p,\text{rel}}_{t,b}-\bs{o}^{g,\text{rel}}_{t,b}\|_2^2 / 0.4^2\Big)$ & 1.0 \\
\\
\multicolumn{3}{l}{\textit{Global tracking only rewards $\mathcal{R}({\bs{s}^{\text{p}}_t}, \bs{s}^{\text{g}}_t)$}} \\
3-point pos (abs.) & $r^{\text{3pt}}_{\text{pos}}(t)=\exp\!\Big(-\tfrac{1}{|\mathcal{B}|}\!\sum_{b\in\mathcal{B}}\|\bs{p}^{p,\text{abs}}_{t,b}-\bs{p}^{g,\text{abs}}_{t,b}\|_2^2 / 0.3^2\Big)$ & 0.5 \\
3-point ori (abs.) & $r^{\text{3pt}}_{\text{ori}}(t)=\exp\!\Big(-\tfrac{1}{|\mathcal{B}|}\!\sum_{b\in\mathcal{B}} \|\bs{o}^{p,\text{abs}}_{t,b}-\bs{o}^{g,\text{abs}}_{t,b}\|_2^2 / 0.4^2\Big)$ & 0.5 \\
\\
\multicolumn{3}{l}{\textit{Penalty terms $\mathcal{P}({\bs{s}^{\text{p}}_t}, \bs{a}_t)$}} \\
% TODO: please add a small gap between rows here -- smaller than \\
Action rate & $r_{\text{act}}(t)=\|\bs{a}_t-\bs{a}_{t-1}\|_2^2$ & -0.1 \\
Joint limit & $r_{\text{jlim}}(t)=\sum_{j} \mathbbm{1}[\bs{q}_{t,j} \notin [\bs{q}_{t,j}^{\text{min}}, \bs{q}_{t,j}^{\text{max}}]]$ & -10.0 \\
Undesired contacts & $r_{\text{contact}}(t)=\sum_{c \notin \{\text{ankles, wrists}\}} \mathbbm{1}[\|\bs{F}_c\|>1.0~\text{N}]$ & -0.1 \\
\bottomrule
\end{tabular}
}
\caption{Reward design details. $\mathcal{B}$: tracked body links.}
\label{tab:rl_rewards}
\end{table}

\subsection{Force perturbation}
\begin{figure}[h]
  \centering
  \includegraphics[width=0.85\linewidth]{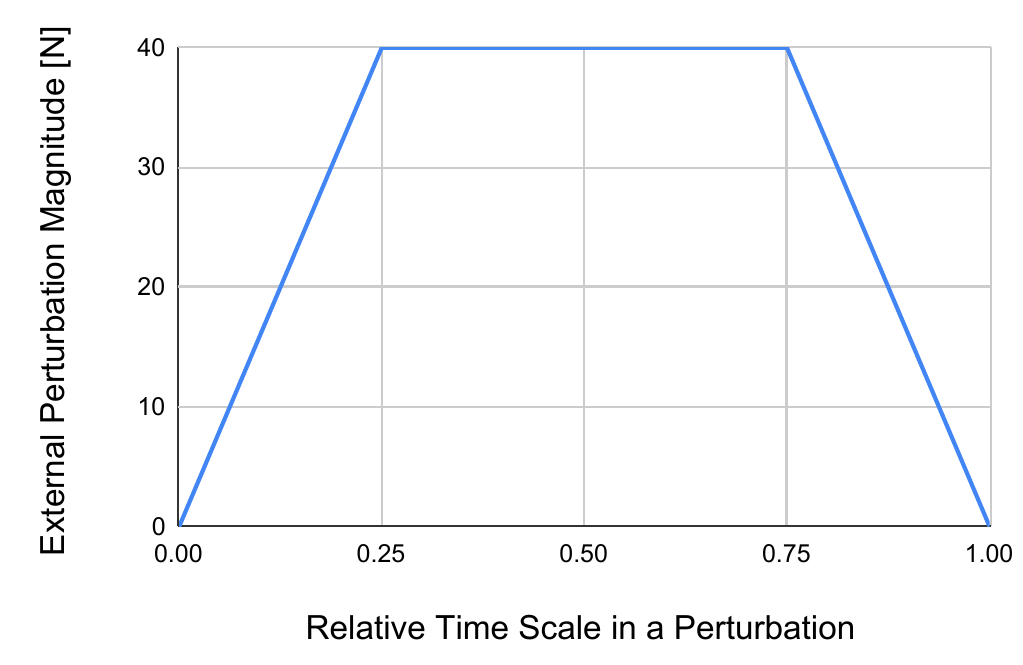}
  \caption{Unified trapezoid profile of external force magnitude schedule within 1-3 seconds perturbation duration.}
  \label{fig:force_mag}
\end{figure}

During training, we apply random force perturbation to the robot's end-effectors. We sample uniformly random force magnitude between 0 and 40 N for each perturbation in a uniformly random direction. For each perturbation, we draw a duration uniformly from 1 to 3 seconds. During the perturbation, the force magnitude changes following the schedule shown in Fig.~\ref{fig:force_mag}. We design this pattern to mimic fundamental contact forces under impedance control, in which the force magnitude gradually increases to its peak at contact and then decreases to zero at break.

\end{document}